\documentclass[10pt,twocolumn,letterpaper]{article}

\usepackage{cvpr}      

\usepackage{graphicx}
\usepackage{amsmath}
\usepackage{amssymb}
\usepackage[title]{appendix}
\usepackage{xspace}
\usepackage{mathtools}
\usepackage[linesnumbered,ruled]{algorithm2e}
\usepackage{bbding}
\usepackage{amsmath}
\usepackage{amsthm}
\usepackage{booktabs}
\usepackage{verbatim}
\usepackage{multirow}
\usepackage{makecell}
\usepackage{times}
\usepackage{epsfig}
\usepackage{pifont}
\usepackage{booktabs}
\usepackage{capt-of}
\usepackage{colortbl}
\usepackage{dsfont}
\usepackage{comment}
\usepackage[dvipsnames]{xcolor}
\usepackage{pifont}
\usepackage{bbm}
\usepackage{url}
\usepackage{nicefrac}
\usepackage{colortbl}
\usepackage[pagebackref,breaklinks,colorlinks,citecolor=cvprblue]{hyperref}

\usepackage[capitalize]{cleveref}
\crefname{section}{Sec.}{Secs.}
\Crefname{section}{Section}{Sections}
\Crefname{table}{Table}{Tables}
\crefname{table}{Tab.}{Tabs.}

\definecolor{cvprblue}{rgb}{0.21,0.49,0.74}

\definecolor{mygray}{gray}{.9}

\definecolor{baselinecolor}{gray}{.9}
\definecolor{darkgreen}{rgb}{0.13, 0.55, 0.13}

\renewcommand{\paragraph}[1]{\vspace{1.25mm}\noindent\textbf{#1}}

\let\originalleft\left
\let\originalright\right
\renewcommand{\left}{\mathopen{}\mathclose\bgroup\originalleft}
\renewcommand{\right}{\aftergroup\egroup\originalright}

\def\paperID{6130} 
\def\confName{CVPR}
\def\confYear{2024}

\begin{document}

\title{Self-supervised Feature Adaptation for 3D Industrial Anomaly Detection}

\author{Yuanpeng Tu$^{1}$\textsuperscript{*} \quad Boshen Zhang$^{2}$\textsuperscript{*}  \quad Liang Liu$^{2}$  \quad Yuxi Li$^{2}$ \quad Xuhai Chen$^{4}$ \\ Jiangning Zhang$^{2}$ \quad Yabiao Wang$^{2\dagger}$ \quad  Chengjie Wang$^{2,3}$ \quad Cai Rong Zhao$^{{1\dagger}}$\\
	$^{1}$Dept. of Electronic and Information Engineering, Tongji Univeristy, Shanghai \\ $^{2}$YouTu Lab, Tencent, Shanghai, \quad  $^{3}$Shanghai Jiao Tong University, \quad  $^{4}$Zhejiang University\\
	{\tt\small \{2030809, zhaocairong\}@tongji.edu.cn}\\
	{\tt\small \{boshenzhang, yukiyxli, leoneliu, vtzhang, caseywang, jasoncjwang\}@tencent.com}}
\maketitle

\maketitle

\begin{abstract}
   Industrial anomaly detection is generally addressed as an unsupervised task that aims at locating defects with only normal training samples. 
   Recently, numerous 2D anomaly detection methods have been proposed and have achieved promising results, however, using only the 2D RGB data as input is not sufficient to identify imperceptible geometric surface anomalies. Hence, in this work, we focus on multi-modal anomaly detection. Specifically, we investigate early multi-modal approaches that attempted to utilize models pre-trained on large-scale visual datasets, i.e., ImageNet, to construct feature databases.  And we empirically find that directly using these pre-trained models is not optimal, it can either fail to detect subtle defects or mistake abnormal features as normal ones. This may be attributed to the domain gap between target industrial data and source data.
   Towards this problem, we propose a Local-to-global Self-supervised Feature Adaptation (LSFA) method to {finetune the adaptors and learn task-oriented representation toward anomaly detection.}
   Both intra-modal adaptation and cross-modal alignment are optimized from a local-to-global perspective in LSFA to ensure the representation quality and consistency in the inference stage.
   Extensive experiments demonstrate that our method not only brings a significant performance boost to feature embedding based approaches, but also outperforms previous State-of-The-Art (SoTA) methods prominently on both MVTec-3D AD and Eyecandies datasets, e.g., LSFA achieves 97.1\% I-AUROC on MVTec-3D, surpass previous SoTA by \textbf{+3.4\%}. 
\end{abstract}

\section{Introduction}
\begin{figure}[t]
  \centering
  \includegraphics[width=\linewidth]{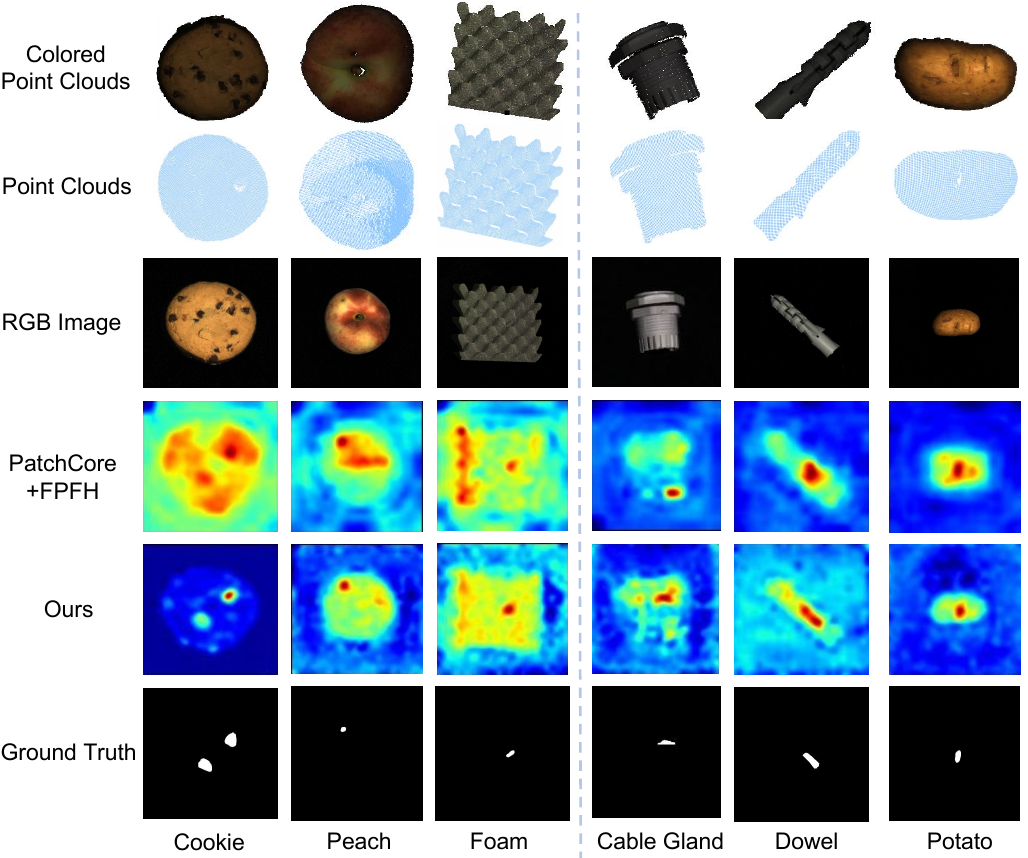}
  \vspace{-5mm}
   \caption{\textbf{Illustrations of MVTec-3D AD dataset}~\cite{mvtec3dad}. The second and third rows are the input point cloud data and RGB data. The fourth and fifth rows are prediction results. Our method can 
   avoid the overestimation issues (as shown left)  
   and produce more accurate results for categories with complex textures (as shown right).}
   \label{fig:motivation}
   \vspace{-3mm}
\end{figure}
    
    Industrial anomaly detection is a widely-explored computer vision task, aiming at detecting unusual image-level/pixel-level patterns in industrial products~\cite{liu2023deep}. 
    Since the lack of anomalous samples in real-world scenarios, current anomaly detection methods usually follow unsupervised paradigm~\cite{patchcore,STFPM,ReverseDistillation,Bergmann_2020_CVPR,Li_2021_CVPR,Hou_2021_ICCV,Wu_2021_ICCV,Ristea_2022_CVPR,Gudovskiy_2022_WACV,wu2022self,yan2021learning}, i.e., training with normal samples but testing on the mixed normal and abnormal samples. Most of the previous methods~\cite{fastflow, STFPM, ReverseDistillation} are designed for 2D images and have achieved great success in 2D anomaly detection. 
    However, in {the scenarios of} industrial inspection, due to lack of depth information, sometimes it is hard to differentiate between subtle surface defect and normal texture with only RGB information (e.g., cookie in Fig~\ref{fig:motivation}.).
    {Therefore, recently there appears new benchmarks~\cite{mvtec3dad,bonfiglioli2022eyecandies} to encourage anomaly detection research in a multi-model view, where the objects are represented with both 2D images and 3D point clouds.} 

    To perform precise anomaly localization, 
    existing 2D anomaly detection approaches can be roughly categorized into two families: reconstruction based and feature embedding based. 
    The former utilizes the characteristic that a generator trained with only normal features cannot successfully reconstruct abnormal features. While the latter aims to model the distribution of normal samples through a well-trained feature extractor, and in inference stage, the out-of-distribution samples are treated as anomalies. The feature embedding based family is more flexible and show promising performance on 2D RGB anomaly detection task. However, simply 
    transferring the 2D feature embedding paradigm into the 3D domain is not easy. Taking the state-of-the-art embedding based method PatchCore~\cite{patchcore} as an example, {when combined} with handcrafted 3D representations (FPFH~\cite{3d-ads}), {it} yields a strong multimodal anomaly detection baseline. However, as shown in Fig.~\ref{fig:motivation}, we experimentally find that the PatchCore+FPFH baseline {shows} two drawbacks, \emph{First,} it is prone to mistake abnormal 
    {regions} as normal ones due to the large discrepancy between {pretrained knowledge} and {industrial scenes} (see the left part in Fig.~\ref{fig:motivation}). \emph{Second,} it sometimes fails to identify small anomaly patterns when it comes to categories with more complex textures, as shown in the right part in Fig.~\ref{fig:motivation}.
    

    To address the aforementioned problems, we resort to a feature adaptation strategy to further enhance the capacity of pre-trained models and learn task-oriented feature descriptors. {\emph{In terms of modality}, color is more effective to identify texture anomalies, while depth information can be helpful to detect geometric deformations in 3D space~\cite{3d-ads}, thus it is more advisable} to leverage both the intra-modal and cross-modal information for adaptation. {On the other hand, \emph{in terms of granularity}}, the {object}-level correspondence between modalities {helps to learn compact representation, while} anomaly detection requires {local sensitivity} to identify subtle {anomalies}~\cite{patchcore}, {hence a multi-grained learning objective is necessary. With these consideration above, }
    we propose a novel \textbf{L}ocal-to-global \textbf{S}elf-supervised multi-modal \textbf{F}eature \textbf{A}daptation framework, named LSFA, to better transfer the {pre-trained} knowledge to {downstream} anomaly detection {task}.
    Specifically, LSFA performs adaptation from two views: intra-modality and cross-modality. {The former adaptation introduces} Intra-modal Feature Compactness (IFC) optimization, where {multi-grained memory banks are applied to learn compact distribution of normal features.} 
    As for the latter one, Cross-modal Local-to-global Consistency (CLC) is {designed to align features from different modality in} both patch-level and {object}-level.
     {With the help of multi-grained information from both modality, model adapted with }
    LSFA yields target-oriented features toward anomaly detection {in 3D space}, {thus} it is {capable of capturing} small anomalies, {while avoiding false positives} (shown in Fig.~\ref{fig:motivation}). 
    For the final inference of anomaly detection, we leverage the fine-tuned features by LSFA to construct memory bank and determine normal/anomaly by computing the feature difference as in~\cite{patchcore}. 
    The effectiveness of LSFA is verified on mainstream benchmarks, including MVTec-3D and Eyecandies. Where LSFA outperforms previous SoTA~\cite{ast} by a large margin, i.e., it obtains 97.1\% (\textbf{+3.4\%}) I-AUROC on MVTec-3D.
    To summarize, the key contributions of this work are as follows:
    
  $\bullet$ We propose LSFA, a novel and effective
  framework towards 3D anomaly detection, it adapts the pre-trained features with local-to-global correspondence between modalities as supervision. 
  It shows significant advantages on mainstream benchmarks and sets the new SOTA.
  
  $\bullet$ In LSFA, Intra-modal Feature Compactness optimization (IFC) is proposed to improve feature compactness from both patch-wise and prototype-wise with dynamic-updated memory banks.

  $\bullet$ In LSFA, Cross-modal Local-to-global Consistency alignment (CLC) is proposed to alleviate cross-modal misalignment with multi-granularity contrastive signals.
  

\begin{figure*}[t]
  \centering
  \includegraphics[width=1.0\linewidth]{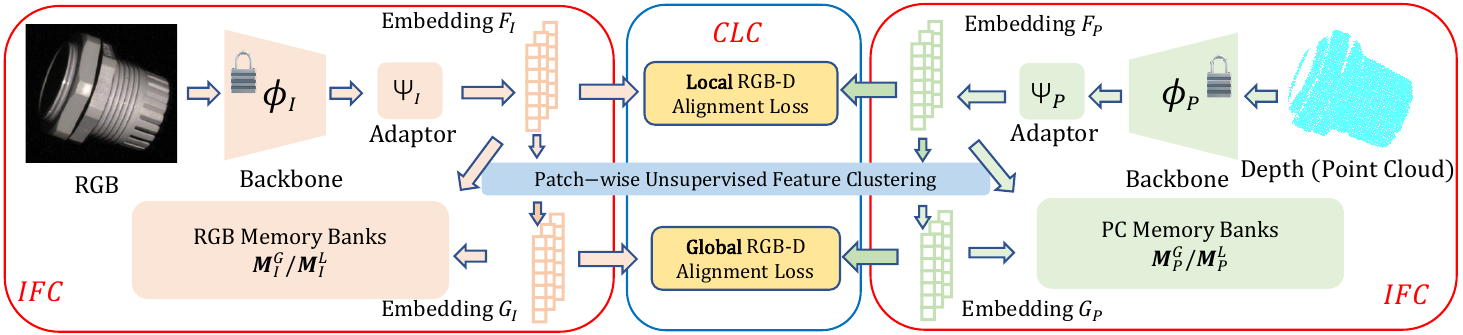}

   \caption{\textbf{The overall pipeline of our method.} The features of two modalities are adapted from two views: Intra-modal Feature Compactness optimization (IFC) and Cross-modal Local-to-global Consistency alignment (CLC). The fine-tuned results of the adaptors are utilized for final defect localization.}
   \label{fig:pipeline}

\end{figure*}

\section{Related Work}
Since our works mainly touch on two aspects of computer vision, namely 2D/3D industrial anomaly detection, we briefly introduce previous traditional 2D/3D industrial anomaly detection approaches respectively in this section.  

\noindent \textbf{2D industrial anomaly detection.} As a binary classification task, unsupervised industrial anomaly detection only trains models with normal samples to distinguish instances sampled from normal/anomaly distribution and localize the anomalous regions, which has drawn extensive attention~\cite{liu2023deep}. Existing methods mainly consist of two categories: reconstruction based and feature embedding based. Among the previous reconstruction based methods, knowledge-distillation based ones~\cite{bergmann2020uninformed, salehi2021multiresolution, ReverseDistillation} assume that there exists a difference between the pre-trained teacher model and the student model in the representation of anomalous patches, where the student model is trained to simulate teacher output for the normal samples during the training process. To prevent negative influence caused by the same filters between teacher and student, ~\cite{ReverseDistillation} integrates a reverse flow paradigm, which can prevent the anomaly gradient propagation to the student as well. ~\cite{bergmann2019mvtec, gong2019memorizing} perform implicit feature modeling and detect defects by comparing the reconstructed images and input ones based on the assumption that anomaly patches cannot be well-recovered.  
    
    Besides these methods, feature embedding based methods recently have achieved state-of-the-art performance by utilizing features extracted from models pre-trained on large-scale external natural image datasets, i.e., ImageNet, and need no further adaption to the data of the target domain. Normalizing flow~\cite{cflow, fastflow} based ones distinguish defects by invertibly transforming normal features into Normal distribution. PaDiM~\cite{padim} takes the correlation between different semantic levels into consideration to extract locally constrained representations and estimate patch-level feature distribution moments. PatchCore~\cite{patchcore} stores normal patch-level features in the memory bank for localizing defects by comparing the target and normal features. CFA~\cite{lee2022cfa} proposes a coupled-hypersphere fine-tuning framework to adapt patch features to the target dataset, thus alleviating the overestimation of the normality of anomaly features.
 
\noindent\textbf{3D industrial anomaly detection.} Different from 2D industrial anomaly detection, 3D industrial anomaly detection identifies anomaly patches by taking both RGB and point cloud samples into consideration. \cite{mvtec3dad} introduces the first public 3D anomaly detection benchmark, MVTec-3D AD for evaluation of methods and builds a baseline approach based on voxel auto-encoder and generative adversarial network, where anomaly defects are located by comparing the voxel-wise difference between the input and reconstructed ones. However, since~\cite{mvtec3dad} lacks of integrating spatially structured information in multi-modal data, only a marginal accuracy boost can be achieved. Inspired by this, \cite{3d-st} proposes a 3D teacher-student framework to extract local-geometry aware descriptors for point clouds and utilizes extra ancillary data for robust pre-training. \cite{3d-ads} firstly explores the appliance of memory bank on this task and utilizes local geometry features extracted from pre-trained models.However, since there is no further adaptation on the target domain of the biased features in~\cite{3d-ads}, there exists a significant performance gap between it and state-of-the-art ones. To address this issue, M3DM~\cite{wang2023multimodal} proposes a multimodal industrial anomaly detection method with hybrid feature fusion to promote interaction between multimodal features. {However, these methods generally perform cross-modal alignment while overlook the importance of intra-modal feature compactness. Therefore, their extracted single-modal features are likely to form a distribution where the anomalous/normal features are difficult to be separated from each other. Such feature distribution limits their ability to effectively integrate information from both modalities as well, leading to inaccurate anomaly detection. Additionally, these methods only consider local-level cross-modal alignment without incorporating global-level alignment of features, which is also crucial for enhancing information interaction between the two modalities.} Motivated by this, we propose a local-to-global self-supervised multimodal adaption method to boost the voxel-level detection performance of feature embedding based approaches from both patch-level and object-level views.
\section{Methodology}
\subsection{Overview}

 \textbf{Framework overview and symbol definition.} In this section, we first {give out the overview of our LSFA} framework. As shown in Fig.~\ref{fig:pipeline}, LSFA takes both point clouds and RGB images in $\mathbb{D}=\{(P_i, I_i)\}_{i=1}^{|\mathbb{D}|}$ as input for joint defect detection, where $P_i \in \mathbb{R}^{N\times3}$ and $I_i \in \mathbb{R}^{H\times W \times 3}$. {For both modality representation $P_i$ and $I_i$, a pretrained feature extractor $\phi_{P}/\phi_{I}$ is applied to obtain modality-specific representation.} Since there exists severe domain bias {between} pre-trained backbones {and downstream detection task}, a vanilla transformer encoder layer~\cite{dosovitskiy2020image} is utilized as the adaptor for these features (note that several other adaptor structures are also investigated in our appendix). {The} adaptors for RGB/3D modalities are denoted as $\Psi_{I}(\cdot)$/$\Psi_{P}(\cdot)$, we propose to perform task-oriented feature adaptation for $\Psi_{I}(\cdot)$/$\Psi_{P}(\cdot)$ from two views: Intra-modal Feature Compactness optimization (IFC) and Cross-modal Local-to-global Consistency alignment (CLC). (1) IFC constructs both global-level and local-level dynamic-updated memory banks for both RGB/3D modality to minimize the distance between normal features from the multi-granularity view, leading to better distinction between normal and abnormal features.
    (2) CLC consists of local-to-global cross-modal alignment modules, which {alleviates} feature misalignment between two modalities and {enhances} the multi-modal information interaction of spatial structures with self-supervised signals. 
    
\noindent\textbf{Inference with adapted representation.} After the adaptation process, since local-sensitive features are more useful for detecting anomaly patterns, the global features are discarded in the final inference stage. For either modal of RGB or Point Cloud, only the local features from adaptor is utilized {to calculate the anomaly score} of each pixel/voxel {through} off-the-shelf PatchCore~\cite{patchcore} algorithm. Finally both anomaly scores from two modalities are averaged as the final anomaly {estimation}.

\subsection{CLC: Cross-modal Local-to-global Consistency Alignment}

    \begin{figure}
  \centering
  \includegraphics[width=0.98\linewidth]{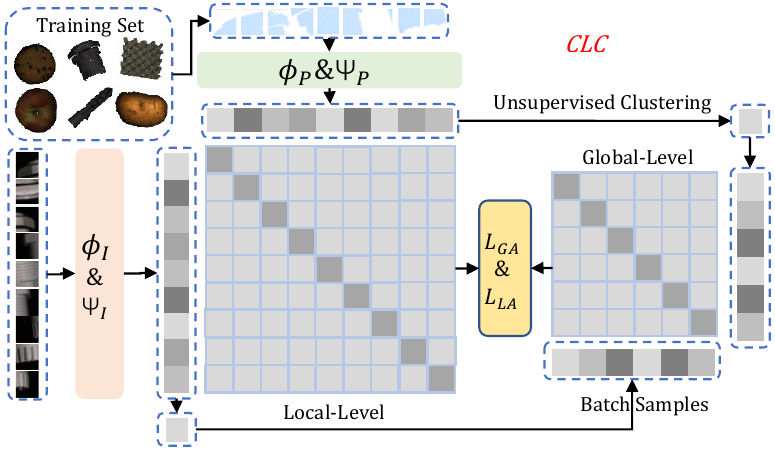}
   \caption{\textbf{The proposed inter-modal local-to-global consistency alignment.} For the local view, similarity of path-wise features in the same/different location of the RGB image and its corresponding 3D point cloud is maximized/minimized to guarantee local-geometry consistency of two modalities. For global view, instance-wise features clustered from patch-wise features are optimized in a similar way.}
   
   \label{fig:IRGC}
   \vspace{-3mm}
\end{figure}

    \textbf{Feature projection.} To extract local-sensitive features for anomaly detection, the ViT~\cite{dosovitskiy2020image} and PointMAE~\cite{pointmae} are utilized as $\phi_{I}/\phi_{P}$. {ViT splits 2D image $I_i$ into $N_m$ patches and extract deep feature for each patch}, {correspondingly, PointMAE group 3D points from $P_i$ into $N_d$ groups and extract group-wise feature}. {To build dense local correspondence between two modality, we remap 3D points into 2D patches via geometric interpolation and projection. Specifically, }
    we denote the deep feature of $i$-th point group as $A_i$ and the group center is denoted as $c_i \in \mathbb{R}^3$, then for each point $p \in \mathbb{R}^3$ in $P_i$, a point-wise deep feature $f_p$ can be obtained via distance-based interpolation: 
\begin{equation}
f_p=\sum_{i=1}^{N_d} \alpha_i A_i, \quad \alpha_i=\frac{\frac{1}{\left\|c_i-p\right\|_2}}{\sum_{k=1}^{N_m} \frac{1}{\left\|c_k-p\right\|_2}}.
\end{equation}
    {Meanwhile, we can verify whether a 3D point $p$ is projected into a 2D patch with camera parameters, thus for each image patch from ViT, we average the feature $f_p$ of all points projected into the same patch as 2D projection of original point-cloud features. By this means, we obtain a 2D patch-wise representation of 3D point features, which shares the same patch number $N_m$ as image features, and the local correspondence is naturally obtained by associating RGB features and projected point features of the same patch. }
     Finally, both patch-wise representation of RGB and point cloud are fed to adaptor $\Psi_{I}(\cdot)$/$\Psi_{P}(\cdot)$ respectively. The adapted features are denoted as $\mathbb{D}_F=\{(F_{P_i}, F_{I_i})\}_{i=1}^{|\mathbb{D}|}$.

\noindent\textbf{Cross-modal local-to-global consistency alignment.} The features of two modalities are aligned in spatial location after the previous step. However, without cross-modal interaction in the adaption process, cross-modal feature misalignment may lead to inferior results when fusing anomalous scores of two modalities during the inference stage. To address this issue, as shown in Fig.~\ref{fig:IRGC}, we perform local-to-global consistency alignment, which can utilize the cross-modal self-supervised signals to enhance feature quality.

Specifically, the adapted patch-wise features for RGB/3D point clouds $\{F_{I_i}, F_{P_i}\}_{i=1}^{N_b}$ are first mapped into the same dimension with two fully-connected layers, denoted as $H_{I}/H_{P}$, where $N_b$ is the batch size. The projected features are denoted as $\{F'_{I_i}, F'_{P_i}\}_{i=1}^{N_b}$, then a patch-wise contrastive loss is calculated to maximize the feature similarity between patches from different modal but the same location, and minimize similarity between patches from different location: 

\begin{equation}
\mathcal{L}_{{LA}}=-\log \left(\frac{\exp \left(\left\langle F_{I_i}^{\prime j}, F_{P_i}^{\prime j}\right\rangle\right)}{\sum_{t=1}^{N_m} \sum_{k=1}^{N_m} \exp \left(\left\langle F_{I_i}^{\prime t}, F_{P_i}^{\prime k}\right\rangle\right)}\right),
\label{eq:infornce}
\end{equation}

\noindent{where $\left<\cdot, \cdot\right>$ denotes the innder production between vectors.} Since Eq.~\ref{eq:infornce} only involves local geometry clues while lacking the interaction of global structural information, we further clustering the local feature $F_{I_i}/F_{P_i}$ to obtain an instance-wise feature $G_{I_i}/G_{P_i}$ with the k-means clustering algorithm.
And then performing a similar operation on this global features, the corresponding \textbf{G}lobal \textbf{A}lignment loss is denoted as $L_{\text {GA}}$.
\begin{equation}
\mathcal{L}_{{GA}}=-\log \left(\frac{\exp \left(\left\langle G_{I_i}^{\prime}, G_{P_i}^{\prime}\right\rangle\right)}{\sum_{t=1}^{N_b} \sum_{x=1}^{N_b} \exp \left(\left\langle G_{I_t}^{\prime}, G_{P_x}^{\prime}\right\rangle\right)}\right).
\label{eq:infornce2}
\end{equation}

Thus the overall loss function for CLC is formulated as:
\begin{equation}
\mathcal{L}_{\text {CLC}}= \mathcal{L}_{\text {LA}} + \mathcal{L}_{\text {GA}}.
\end{equation}

\subsection{IFC: Intra-modal Feature Compactness Optimization}

    The proposed intra-modal feature compactness optimization strategy aims at helping models generate more {compact representation} for normal samples, thus making models more sensitive to {anomaly} patterns.

\noindent\textbf{Local-to-global compactness optimization.} Since there exists severe domain bias for the pre-trained models without adaptation, the extracted features are likely to form a distribution where the anomalous/normal features are difficult to be separated from each other. Consequently, previous feature embedding based methods~\cite{3d-ads} are inevitably prone to {mistake anomalies as normal areas}. Motivated by this, as shown in Fig.~\ref{fig:memorybank}, we design a dynamic-updated memory-bank {in both local and global level to guide compactness optimization}. 
        
     {Since the optimization is conducted within each modality, here} we take RGB feature as an example {and the point-cloud feature is processed in a similar manner. Concretely,} we denote the memory bank consisting of patch-level RGB features as ${M_{I}^{L}}$ {with length $|M_{I}^{L}| = n_{I}^{L}$}. The j-th patch-level feature $F_{I_i}^{j}$ of $I_i$ in batch $\{F_{I_i}\}_{i=1}^{N_b}$ is utilized for nearest neighbor searching in ${M_{I}^{L}}$, where $N_b$ is the batch size. A mean squared error loss is utilized to minimize the discrepancy between $F_{I_i}^{j}$ and its corresponding nearest item in ${M_{I}^{L}}$. Thus the \textbf{L}ocal patch-level \textbf{C}ompactness $\mathcal{L}_{\text {LC}}$ loss can be derived as follows:

\begin{figure}
  \centering
  \includegraphics[width=0.95\linewidth]{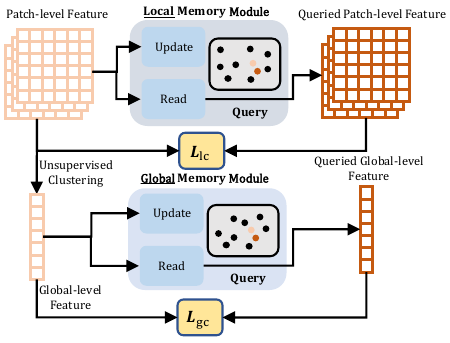}
  \vspace{-2mm}
   \caption{\textbf{The proposed local-to-global compactness optimization strategy}, where both prototype-wise global-level and patch-wise local-level memory banks are involved.}
   \label{fig:memorybank}

\end{figure}

\begin{equation}
\mathcal{L}_{\text {LC}}= \sum_{i=1}^{N_b} \sum_{j=1}^{N_m}
\min _{\mathrm{Q} \in \mathrm{M}_{\mathrm{I}}^{\mathrm{L}}} \left\|F_{I_i}^j-Q\right\|_2.
\end{equation}
        Where $N_m$ is the patch number. Furthermore, to enhance the compactness of features for each category, a global compactness loss is designed to simultaneously optimize the global feature $G_{I_i}$. 
        Denote the memory bank consisting of global RGB features with length $n_{I}^{G}$ as ${M_{I}^{G}}$. A similar nearest neighbor search operation is performed for $G_{I_i}$ and ${M_{I}^{G}}$ to enhance sensitivity against anomalies from the global view. Therefore, the \textbf{G}lobal \textbf{C}ompactness loss $\mathcal{L}_{\text {GC}}$ is:
\begin{equation}
\mathcal{L}_{\text {GC}}= \sum_{i=1}^{N_b} \min _{\mathrm{Q} \in \mathrm{M}_{\mathrm{I}}^{\mathrm{G}}} \left\|G_{I_i}^j-Q\right\|_2.
\end{equation}

        After each iteration, the local-level features and global-level features of current batch samples are enqueued into ${M_{I}^{L}}$/${M_{I}^{G}}$ respectively, which can be derived as:
\begin{equation}
\left\{\begin{array}{l}
M^L_I = M^L_I \cup \{F^j_{I_i} | j \in [1,N_m], i \in [1,{N_b}]\} \\
M^G_I = M^G_I \cup \{G_{I_i} | i \in [1,{N_b}]\}.
\end{array}\right.
\end{equation}

\begin{algorithm}

\caption{Training for the proposed LSFA.}\label{alg:LSFA}
\KwIn {Memory banks $\{M_{I}^{G}, {M_{I}^{L}}$, ${M_{P}^{G}}, M_{P}^{L}\}$, adaptors $\{\Psi_{I},\Psi_{P}\}$, linear projection layer $\{H_{I},H_{P}\}$, training set features $\{F_{I}, F_{P}\}$.}

\KwOut {Parameters of adaptors {$\{\Theta_{I}, \Theta_{P}\}$.}}

Initialize $M_{I}^{G}, M_{I}^{L}, M_{P}^{G}, M_{P}^{L}$.\\
\For{${F_{I_i}},F_{P_i}\in \mathbb{D}_F$}{
{   \small
    $F'_{I_i} {\longleftarrow} H_{I}(F_{I_i})$; $F'_{P_i} {\longleftarrow} H_{P}(F_{P_i})$ 
    \tcc{Inter-modal Local-to-global Consistency Alignment}  
    $ \Theta_{I},\Theta_{P} \stackrel{optim}{\longleftarrow} \mathcal{L}_{\text {CLC}}(F'_{I_i}; F'_{P_i}; \Theta_{P}; \Theta_{I})$ 
    \tcc{Cross-modal Feature Compactness Optimization}  
    $ \Theta_{I} \stackrel{optim}{\longleftarrow} \mathcal{L}_{\text {IFC}}(F_{I_i};M_{I}^{G}; M_{I}^{L} ;\Theta_{I})$ 
    $ \Theta_{P} \stackrel{optim}{\longleftarrow} \mathcal{L}_{\text {IFC}}(F_{P_i};M_{P}^{G}; M_{P}^{L};\Theta_{P})$ 
    \tcc{Update Memory Banks}  
    $ M_{I}^{G}, M_{I}^{L} \stackrel{update}{\longleftarrow} F_{I_i}$; $ M_{P}^{G}, M_{P}^{L} \stackrel{update}{\longleftarrow} F_{P_i}$
    }
}
\end{algorithm}

Meanwhile, the least recently appended features with the same length as the enqueued features will be popped out from ${M_{I}^{L}}$/${M_{I}^{G}}$ to keep the features in banks up-to-date when the length of ${M_{I}^{L}}$/${M_{I}^{G}}$ is larger than $n_{I}^{G}$/$n_{I}^{L}$. Similar global and local compactness optimization operations are performed for the point cloud features $\{F_{P_i}\}_{i=1}^{N_b}$ as well, where the global and local memory bank sizes of point cloud features are the same as RGB modality.

Consequently, the loss function of the proposed IFC can be summarized as:

\begin{equation}
\mathcal{L}_{\text {IFC}}=\mathcal{L}_{\text {LC}} + \mathcal{L}_{\text {GC}}.
\end{equation}
Therefore to summarize, the overall training loss for the proposed LSFA is derived as:
\begin{equation}
\mathcal{L}_{\text {LSFA}}= \mathcal{L}_{\text {IFC}} + \lambda \mathcal{L}_{\text {CLC}}.
\end{equation}
Where $\lambda$ is a balancing hyper-parameter.

\begin{table*}
  \centering
  \scriptsize
  \resizebox{\linewidth}{!}{
  \renewcommand\arraystretch{0.8}
  \begin{tabular}{@{}cl|m{0.8cm}<{\centering}m{0.8cm}<{\centering}m{0.8cm}<{\centering}m{0.8cm}<{\centering}m{0.8cm}<{\centering}m{0.8cm}<{\centering}m{0.8cm}<{\centering}m{0.8cm}<{\centering}m{0.8cm}<{\centering}m{0.8cm}<{\centering}|m{0.8cm}<{\centering}}
    \toprule
    & Method & Bagel & Cable Gland & Carrot & Cookie & Dowel & Foam & Peach & Potato & Rope & Tire & Mean\\
    \midrule
     \multirow{10}*{\rotatebox{90}{3D}}&Depth GAN\cite{mvtec3dad} & 0.530 & 0.376 & 0.607 & 0.603 & 0.497 & 0.484 & 0.595 & 0.489 & 0.536 & 0.521 & 0.523\\
     ~&Depth AE\cite{mvtec3dad} & 0.468 & \underline{0.731} & 0.497 & 0.673 & 0.534 & 0.417 & 0.485 & 0.549 & 0.564 & 0.546 & 0.546\\
     ~&Depth VM\cite{mvtec3dad} & 0.510 & 0.542 & 0.469 & 0.576 & 0.609 & 0.699 & 0.450 & 0.419 & 0.668 & 0.520 & 0.546\\
     ~&Voxel GAN\cite{mvtec3dad} & 0.383 & 0.623 & 0.474 & 0.639 & 0.564 & 0.409 & 0.617 & 0.427 & 0.663 & 0.577 & 0.537\\
     ~&Voxel AE\cite{mvtec3dad} & 0.693 & 0.425 & 0.515&  0.790 & 0.494 & 0.558 & 0.537 & 0.484 & 0.639 & 0.583 & 0.571\\
     ~&Voxel VM\cite{mvtec3dad} & 0.750  & \textbf{0.747} & 0.613 & 0.738 & 0.823 & 0.693 & 0.679 & 0.652 & 0.609 & 0.690 & 0.699\\
     ~&3D-ST\cite{3d-st} & 0.862& 0.484 & 0.832 & 0.894 & 0.848 & 0.663 & 0.763 & 0.687 & \underline{0.958} & 0.486 & 0.748 \\
     ~&FPFH\cite{3d-ads} & 0.825 & 0.551 & 0.952 & 0.797 & 0.883 & 0.582 & 0.758 & 0.889 & 0.929 & 0.653 & 0.782\\
     ~&AST\cite{ast} & {0.881} & 0.576 & \underline{ 0.965} & 0.957 & 0.679 & \underline{0.797} & \textbf{0.990} & 0.915 & 0.956 & 0.611 & 0.833\\

     ~&FPFH*/M3DM\cite{wang2023multimodal} & \underline{0.941} & 0.651 & \underline{ 0.965} & \underline{ 0.969} & \underline{ 0.905} & {0.760} & {0.880} & \textbf{0.974}& 0.926 & \underline{ 0.765} & \underline{0.874}\\
     \rowcolor{mygray}
     ~&LSFA(Ours) & \textbf{0.986 } & 0.669  & \textbf{0.973 } & \textbf{0.990 } & \textbf{ 0.950} & \textbf{0.802} & \underline{0.961} & \underline{0.964}&\textbf{0.967}  & \textbf{ 0.944} & \textbf{0.921}\\
    \midrule

      \multirow{7}*{\rotatebox{90}{RGB}}&DifferNet\cite{differnet} & 0.859 & 0.703 & 0.643 & 0.435 & 0.797 & 0.790 & 0.787 & {0.643} & 0.715 & 0.590 & 0.696\\
     &PADiM\cite{padim} & \textbf{ 0.975} & 0.775 & 0.698 & 0.582 & 0.959 & 0.663 & 0.858 & 0.535 & 0.832 & 0.760 & 0.764\\
     &PatchCore\cite{patchcore} & 0.876 & 0.880 & 0.791 & 0.682 & 0.912 & 0.701 & 0.695 & 0.618 & 0.841 & 0.702 & 0.770\\
     &STFPM\cite{STFPM} & 0.930 & 0.847 & {0.890} & 0.575 & 0.947 & 0.766 & 0.710 & 0.598 & 0.965 & 0.701 & 0.793\\
     &CS-Flow\cite{cflow} & 0.941 & \textbf{ 0.930} & 0.827 & \underline{0.795} & \textbf{ 0.990} & \underline{0.886} & 0.731 & 0.471 &  \underline{0.986} & 0.745 & 0.830\\
     &AST\cite{ast} & { 0.947} & \underline{0.928} & 0.851 & \textbf{ 0.825} & \underline{0.981} & \textbf{ 0.951} & {0.895} & 0.613 & \textbf{ 0.992}& \textbf{ 0.821} & \textbf{ 0.880}\\
     &PatchCore*/M3DM\cite{wang2023multimodal} & 0.944 & 0.918 & \underline{0.896} & 0.749 & 0.959 & 0.767 & \underline{0.919} & \underline{ 0.648} & 0.938 & {0.767} & {0.850}\\
     \rowcolor{mygray}
          ~&LSFA(Ours) & \underline{0.951} & 0.920 & \textbf{0.911} & 0.762 & 0.961 & 0.770 & \textbf{ 0.930} & \textbf{ 0.675} & 0.938 & \underline{0.787} & \underline{0.861}\\
    \midrule
     \multirow{10}*{\rotatebox{90}{RGB + 3D}}&Depth GAN\cite{mvtec3dad} & 0.538 & 0.372& 0.580& 0.603& 0.430& 0.534& 0.642& 0.601& 0.443& 0.577& 0.532\\
     &Depth AE\cite{mvtec3dad} & 0.648 & 0.502 & 0.650 & 0.488& 0.805 & 0.522& 0.712 & 0.529 & 0.540 & 0.552 & 0.595\\
     &Depth VM\cite{mvtec3dad} & 0.513& 0.551& 0.477 & 0.581 & 0.617 & 0.716 & 0.450 & 0.421& 0.598& 0.623& 0.555\\
     &Voxel GAN\cite{mvtec3dad} & 0.680& 0.324& 0.565 & 0.399& 0.497& 0.482& 0.566& 0.579& 0.601& 0.482& 0.517\\
     &Voxel AE\cite{mvtec3dad} & 0.510& 0.540 & 0.384& 0.693& 0.446& 0.632& 0.550& 0.494& 0.721& 0.413& 0.538\\
     &Voxel VM\cite{mvtec3dad} & 0.553 & 0.772& 0.484& 0.701& 0.751& 0.578& 0.480& 0.466& 0.689& 0.611& 0.609\\
     &3D-ST\cite{3d-st} & 0.950 & 0.483 & \textbf{ 0.986} & 0.921 & 0.905 & 0.632 & 0.945 & \textbf{ 0.988} & {0.976} & 0.542 & 0.833\\
     &PatchCore + FPFH\cite{3d-ads} & 0.918 & 0.748 & 0.967 & 0.883 & {0.932} & 0.582 & 0.896 & 0.912 & 0.921 & \underline{ 0.886} & 0.865\\
     &AST\cite{ast}  &  {0.983} & {0.873} & {0.976} & {0.971} & {0.932} & {0.885} & \underline{ 0.974} & \underline{0.981} & \textbf{ 1.000} & 0.797 & {0.937} \\
    &PatchCore*+FPFH*\cite{3d-ads} & { 0.981} & { 0.831} & 0.980 & \underline{ 0.985} & \underline{ 0.960} & { 0.905} & {0.936} & 0.964& 0.967 & {0.780} & { 0.929}\\
    &M3DM\cite{wang2023multimodal}     & \underline{0.994} & \underline{0.909} & 0.972 & 0.976 &\underline{0.960} & \underline{0.942} &0.973 & 0.899 &0.972& 0.850 & \underline{0.945} \\
    \rowcolor{mygray}
    ~&LSFA(Ours) & \textbf{1.000 } & \textbf{0.939}  & \underline{0.982 } & \textbf{ 0.989} & \textbf{ 0.961} & \textbf{0.951} & \textbf{0.983} & 0.962&\underline{0.989}  & \textbf{0.951 } & \textbf{0.971}\\
    \bottomrule
  \end{tabular}}
  \vspace{-2mm}
  \caption{\textbf{I-AUROC for anomaly detection of all categories of MVTec-3D AD.} '*' denotes replacing its features with the same pre-trained features as LSFA for PatchCore. Results with confidence intervals of LSFA are shown in the supplementary material.
  }
  \label{tab:iaucroc}
\vspace{-2mm}
\end{table*}

\subsection{Defect Localization}

 Since LSFA is designed for adapting pre-trained features to estimate anomaly patterns better. We utilize the pre-trained backbones and the adaptors for final feature extraction. The adapted features of two modalities are respectively fed into the off-the-shelf feature embedding based method PatchCore~\cite{patchcore}. The anomaly scores of two modalities are averaged as the final anomaly score for each pixel/voxel to evaluate the effectiveness on anomaly detection. The overall pseudo-code of LSFA can be found in Algorithm~\ref{alg:LSFA}.


\noindent\textbf{Discussion. }{Since the framework of LSFA is similar to M3DM~\cite{wang2023multimodal}, here we discuss their difference in detail. First, rather than introducing extra modules for feature fusion in ~\cite{wang2023multimodal}, we only perform feature adaptation for each modality and needs no extra memory bank, thus introducing no extra time and memory cost for inference. Moreover, M3DM overlooks the importance of object-level feature alignment to accurate anomaly detection. And our LSFA performs cross-modal feature alignment from both object-level and patch-level views to fully enhance the consistency and interaction of cross-modal discriminative information, thus demonstrating much superior performance to it. Finally, LSFA takes the intra-modal feature compactness into consideration, which is ignored in M3DM as well. Specifically, similar to cross-modal alignment, the intra-modal feature compactness optimization is also conducted from both patch-level and object-level perspectives to alleviate the influence of domain bias of pre-trained features and obtain high-quality single-modal features.}

\section{Experiments}
\subsection{Experimental Details}
\textbf{Dataset.} To verify the effectiveness of LSFA, we conduct experiments on two 3D industrial anomaly detection datasets: MVTec-3D AD~\cite{mvtec3dad} and Eyecandies~\cite{bonfiglioli2022eyecandies}. Details of the datasets are discussed in our supplementary material. 



\noindent\textbf{Implementation details.} For the feature extractors of the RGB modality, a ViT-B/8~\cite{dosovitskiy2020image} pre-trained on ImageNet~\cite{deng2009imagenet} with DINO~\cite{caron2021emerging} is adopted for a balance of efficiency and performance. The 768-dim output of the final layer is used and then pooled into 56$\times$56 for subsequent training. For the feature extraction of 3D modality, a point transformer~\cite{pang2022masked} pre-trained on ShapeNet~\cite{chang2015shapenet} dataset is utilized and the outputs from 3/7/11 layer are concatenated to fuse multi-scale information for further fine-tuning. Similar to patches in ViT, the point transformer clusters point clouds into multiple local groups and these groups have their corresponding center points for position and neighbor numbers for group size. For data processing, the background area of depth and RGB images is removed by estimating the background plane of depth images with RANSAC~\cite{fischler1981random}, where points within 5$\times$$10^{-2}$ are ignored to accelerate the feature extraction and meanwhile alleviate the influence of background. Finally, the RGB and point cloud tensor are both resized to 224 $\times$ 224 to be consistent with the input size. The projected features for point clouds and RGB samples in CLC are 512-dim. The AdamW optimizer is used and its learning rate is set as 2$\times$$10^{-3}$ with cosine warm-up. The batch size $N_b$ for adaptation is set as 8.

\begin{table*}
\vspace{-7mm}
  \centering
  \scriptsize
  \begin{tabular}{@{}cl|m{0.8cm}<{\centering}m{0.8cm}<{\centering}m{0.8cm}<{\centering}m{0.8cm}<{\centering}m{0.8cm}<{\centering}m{0.8cm}<{\centering}m{0.8cm}<{\centering}m{0.8cm}<{\centering}m{0.8cm}<{\centering}m{0.8cm}<{\centering}|m{0.8cm}<{\centering}}
    \toprule
    & Method & Bagel & Cable Gland & Carrot & Cookie & Dowel & Foam & Peach & Potato & Rope & Tire & Mean\\
    \midrule
     \multirow{8}*{\rotatebox{90}{3D}}&Depth GAN\cite{mvtec3dad} & 0.111& 0.072& 0.212& 0.174 & 0.160 & 0.128 & 0.003 & 0.042 & 0.446 & 0.075 & 0.143\\
     ~&Depth AE\cite{mvtec3dad} & 0.147 & 0.069 & 0.293 & 0.217 & 0.207 & 0.181 & 0.164 & 0.066 & 0.545 & 0.142 & 0.203 \\
     ~&Depth VM\cite{mvtec3dad} & 0.280 & 0.374 & 0.243 & 0.526 & 0.485 & 0.314 & 0.199 & 0.388 & 0.543 & 0.385 & 0.374\\
     ~& Voxel GAN\cite{mvtec3dad} & 0.440 & 0.453 & 0.875 & 0.755 & 0.782 & 0.378 & 0.392 & 0.639 & 0.775 & 0.389 & 0.583 \\
     ~&Voxel AE\cite{mvtec3dad} & 0.260 & 0.341 & 0.581 & 0.351 & 0.502 & 0.234 & 0.351 & 0.658 & 0.015 & 0.185 & 0.348 \\
     ~&Voxel VM\cite{mvtec3dad} & 0.453 & 0.343 & 0.521 & 0.697 & 0.680 & 0.284 & 0.349 & 0.634 & 0.616 & 0.346 & 0.492\\
     ~&FPFH\cite{3d-ads} & \underline{ 0.973} & \underline{ 0.879} & \textbf{ 0.982} & \underline{ 0.906} & \underline{ 0.892} & {0.735} &\underline{0.977} & \underline{ 0.982} & \underline{ 0.956} & \underline{ 0.961} & \underline{ 0.924}\\
     &FPFH*/M3DM\cite{wang2023multimodal} & {0.943} & {0.818} & {0.977} & {0.882} & {0.881} & \underline{ 0.743} & {0.958} & {0.974} & {0.950} & {0.929} & {0.906} \\
     \rowcolor{mygray}
     ~&LSFA(Ours) & \textbf{0.974} & \textbf{0.887}  & \underline{0.981 } & \textbf{ 0.921} & \textbf{ 0.901} & \textbf{0.773} & \textbf{0.982} & \textbf{0.983}& \textbf{0.959} & \textbf{ 0.981} & \textbf{0.934}\\
    \midrule
     \multirow{3}*{\rotatebox{90}{RGB}}&PatchCore\cite{patchcore} & {0.901} & {0.949} & {0.928} & {0.877} & {0.892} & {0.563} & {0.904} & {0.932} & {0.908} & {0.906} & {0.876}\\
     & PatchCore*/M3DM\cite{wang2023multimodal} & \underline{ 0.952} & \underline{ 0.972} & \textbf{ 0.973} & \underline{ 0.891} & \underline{ 0.932} & \underline{ 0.843} & \textbf{ 0.970} & \underline{ 0.956} & \underline{ 0.968} & \textbf{ 0.966} & \underline{ 0.942} \\
     \rowcolor{mygray}
    ~&LSFA(Ours) & \textbf{0.957} & \textbf{0.976}  & \underline{0.970 } & \textbf{ 0.912} & \textbf{0.934 } & \textbf{0.851} & \underline{0.960} &  \textbf{0.957} & \textbf{0.970 } & \underline{0.961} & \textbf{0.945}\\
    \midrule
     \multirow{9}*{\rotatebox{90}{RGB + 3D}}&Depth GAN\cite{mvtec3dad} & 0.421 & 0.422& 0.778 & 0.696 & 0.494 & 0.252 & 0.285 & 0.362 & 0.402 & 0.631 & 0.474 \\
     &Depth AE\cite{mvtec3dad} & 0.432 & 0.158 & 0.808 & 0.491 & 0.841 & 0.406 & 0.262 & 0.216 & 0.716 & 0.478 & 0.481 \\
     &Depth VM\cite{mvtec3dad} & 0.388 & 0.321 & 0.194 & 0.570 & 0.408 & 0.282 & 0.244 & 0.349 & 0.268 & 0.331 & 0.335\\
     &Voxel GAN\cite{mvtec3dad} & 0.664 & 0.620 & 0.766 & 0.740 & 0.783 & 0.332 & 0.582 & 0.790 & 0.633 & 0.483 & 0.639 \\
     &Voxel AE\cite{mvtec3dad} & 0.467 & 0.750 & 0.808 & 0.550 & 0.765 & 0.473 & 0.721 & 0.918 & 0.019 & 0.170 & 0.564\\
     &Voxel VM\cite{mvtec3dad} & 0.510 & 0.331 & 0.413 & 0.715 & 0.680 & 0.279 & 0.300 & 0.507 & 0.611 & 0.366 & 0.471\\
     &3D-ST\cite{3d-st} & 0.950 & 0.483 & \textbf{ 0.986} & 0.921 & 0.905 & 0.632 & 0.945 & \textbf{ 0.988} & \textbf{ 0.976} & 0.542 & 0.833\\
     &PatchCore + FPFH\cite{3d-ads} & \underline{ 0.976} & \underline{0.969} & {0.979} & \textbf{ 0.973} & \underline{0.933} & {0.888} & {0.975} & {0.981} & 0.950 & {0.971} & {0.959} \\
     &PacthCore*+FPFH*\cite{3d-ads} & {0.968} & { 0.925} & {0.979} & {0.914} & { 0.909} & \textbf{ 0.948} & { 0.975} & 0.976  & {0.967} & { 0.965} & { 0.953}\\
     &M3DM~\cite{wang2023multimodal}  & {0.970}  &\underline{0.971}  &{0.979} &\underline{0.950}  & \textbf{0.941}   &0.932  & \underline{0.977}    &0.971  & 0.971 & \underline{0.975}  & \underline{0.964}  \\
     \rowcolor{mygray}
    ~&LSFA(Ours) & \textbf{0.986} & \textbf{0.974}  & \underline{0.981 } & {0.946 } & {0.925 } & \underline{0.941} & \textbf{0.983} &\underline{0.983} & \underline{0.974} & \textbf{0.983 } & \textbf{0.968}\\     
    \bottomrule
  \end{tabular}
  \vspace{-1.1em}
  \caption{\textbf{AUPRO for anomaly segmentation of all categories of MVTec-3D.} '*' denotes replacing its features with the same pre-trained features as LSFA for PatchCore. Results with confidence intervals of LSFA are shown in the supplementary material.
  }
  \label{tab:aupro}
  \vspace{-1mm}
\end{table*}

\noindent\textbf{Evaluation metrics.} Following the standard evaluation protocol of MVTec-3D and Eyecandies, we use Image-level ROCAUC (I-AUROC), pixel-wise AUROC (P-AUROC), and the overlap of each region (AUPRO) to present the anomaly detection performance. The I-AUROC/P-AUROC is defined as the area under the receiver operator curve of image-level/pixel-level predictions, while AUPRO denotes the average relative overlap of binary prediction with each connected component of ground truth labels.  
\subsection{Comparison on 3D AD Benchmark}
\vspace{-2mm}
    To comprehensively evaluate the effectiveness of our method, we first conduct experiments on both 3D/RGB/3D+RGB modality on MVTec-3D AD. Tab.~\ref{tab:iaucroc} and Tab. ~\ref{tab:aupro} present the comparison results of I-AUROC and AUPRO, the methods are grouped by modality (we also report P-AUROC in the supplementary material). 1) For the I-AUROC metric, our method can not only bring a significant boost to the baseline method on both single-modality benchmarks but also multi-modality combined ones, especially for the challenging categories, e.g., cable gland and tire. The single-modality results demonstrate that our intra-modal feature compactness optimization effectively improves the feature quality, thus benefiting the anomaly localization in the inference process. Moreover, our method significantly outperforms all previous methods regarding the average of all classes by a large margin of 4.7\% for 3D, and 4.2\% for the combination. A new state-of-the-art performance is achieved in 17 of all 30 cases for all the individual classes and data modalities. 2) For the AUPRO metric, LSFA can also achieve consistently higher scores than all previous methods for anomaly segmentation, demonstrating that our method is better at mining localized and detailed clues to discover crucial unexpected patterns. Besides MVTec-3D AD, we further perform a detailed evaluation on the latest large-scale 3D AD dataset Eyecandies. The corresponding results are shown in the supplementary material, where our method obtains the best results and significantly outperforms all the previous approaches, achieving the average I-AUROC/AUPRO \textcolor{black}{of 87.5\% and 97.8\%} respectively for RGB modality.

\subsection{Ablation Study}

To study the influence of each component within the proposed LSFA, we conduct ablation analysis on MVTec-3D.

\begin{figure}
  \centering
  \includegraphics[width=0.95\linewidth]{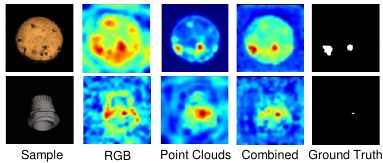}
  \vspace{-.7em}
   \caption{Qualitative results of RGB/D modality. 
   }
   \label{fig:ablationmemorybank}
   \vspace{-2mm}
\end{figure}

\begin{figure}
  \centering
  \includegraphics[width=1.0\linewidth]{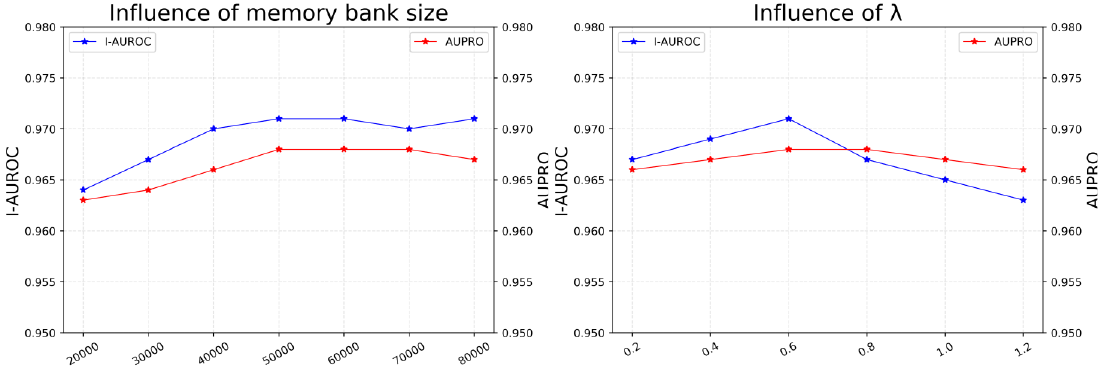}
  \vspace{-1.8em}
   \caption{Investigation on the influence of memory bank size $n_{I}^{L}$ (left) and balancing hyber-parameter $\lambda$ (right). 
   }
   \label{fig:ablationparameter}

\end{figure}

\begin{table*}[t]
\begin{minipage}{0.66\linewidth}
\begin{minipage}{0.47\linewidth}
   \footnotesize
   \vspace{-1mm}
   \centering
   \resizebox{\linewidth}{!}{
  \begin{tabular}{cc|ccc}
    \toprule
    \multicolumn{2}{c|}{Component} & \multirow{2}*{I-AUROC} &  \multirow{2}*{AUPRO} & \multirow{2}*{P-AUROC}\\
    \cline{1-2} $L_{\text {GA}}$ & $L_{\text{LA}}$ &   &  &    \\
    \midrule 
    {\XSolidBrush} & {\XSolidBrush} &  0.929   &0.953  & 0.987 \\
    {\XSolidBrush} & {\Checkmark} & 0.949   & 0.961   & 0.989    \\
    {\Checkmark} & {\XSolidBrush} & 0.952  & 0.961    & 0.990    \\
     {\Checkmark}& {\Checkmark} & \textbf{0.959}  & \textbf{0.964}  & \textbf{0.992}    \\
    \bottomrule

  \end{tabular}}\vspace{-2mm}
     \caption{Investigation on the loss functions within CLC.}\vspace{-2mm}
    
  \label{tab:ablation_componentCLC}
\end{minipage}
\hspace{1mm}
\begin{minipage}{0.47\linewidth}
   \footnotesize
   \centering
   \resizebox{\linewidth}{!}{
  \begin{tabular}{cc|ccc}
    \toprule
    \multicolumn{2}{c|}{Component} & \multirow{2}*{I-AUROC} &  \multirow{2}*{AUPRO} & \multirow{2}*{P-AUROC}\\
    \cline{1-2} IFC & CLC &   &  &    \\
    \midrule 
    {\XSolidBrush} & {\XSolidBrush} & 0.929   & 0.953    & 0.987  \\
    {\XSolidBrush} & {\Checkmark} & 0.957  &0.963    &0.990     \\
    {\Checkmark} & {\XSolidBrush} & 0.959  &0.964    &0.992     \\
     {\Checkmark}& {\Checkmark} & \textbf{0.971}  & \textbf{0.968}  & \textbf{0.993}    \\
    \bottomrule

  \end{tabular}}\vspace{-2mm}
     \caption{Ablation results for two components in LSFA, i.e., IFC and CLC.}\vspace{-2mm}
  \label{tab:ablation_componentall}
\end{minipage}
\end{minipage}
\begin{minipage}{0.33\linewidth}

   \footnotesize
  \resizebox{\linewidth}{!}{
  \begin{tabular}{cc|ccc}
    \toprule
    \multicolumn{2}{c|}{Component} & \multirow{2}*{I-AUROC} &  \multirow{2}*{AUPRO} & \multirow{2}*{P-AUROC}\\
    \cline{1-2}  $\mathcal{L}_{\text {GC}}$ & $\mathcal{L}_{\text {LC}}$  &  &    \\
    \midrule 
    {\XSolidBrush} & {\XSolidBrush} & 0.929   & 0.953    &0.987   \\
    {\Checkmark} & {\XSolidBrush} & 0.950  & 0.960   & 0.988    \\
    {\XSolidBrush} & {\Checkmark} & 0.952  & 0.960   & 0.989\\
    {\Checkmark} & {\Checkmark} & \textbf{0.957}  & \textbf{0.963}  & \textbf{0.990}    \\
    \bottomrule
  \end{tabular}}
  \vspace{-2mm}
  \caption{Investigation on the loss functions within IFC. 
  }
  
  \label{tab:ablation_componentIFC}
\end{minipage}
 \end{table*}

\begin{table*}[t]
\begin{minipage}{0.66\linewidth}
\begin{minipage}{0.47\linewidth}
   \scriptsize
   \vspace{-1mm}
   \centering
  \setlength{\tabcolsep}{0.7mm}{
  \begin{tabular}{c|ccc}
    \toprule
    \multicolumn{1}{c|}{Structure $\Psi_I/\Psi_P$} & \multirow{1}*{I-AUROC} &  \multirow{1}*{AUPRO} & \multirow{1}*{P-AUROC}\\
    \midrule 
    Linear projection &0.953  & 0.959 &0.989 \\
    Single encoder layer & \textbf{0.974}   & \textbf{0.968}   & \textbf{0.993} \\
    Two encoder layers  & 0.954 & 0.963 &0.984 \\
    1$\times$1 Convolution & 0.951  & 0.962 &0.986 \\

    \bottomrule
  \end{tabular}
  }\vspace{-2mm}
  \caption{Investigation on the structure of $\Psi_I/\Psi_P$.}
  \label{tab:structure}
\end{minipage}
\hspace{1mm}
\begin{minipage}{0.47\linewidth}
   \footnotesize
   \centering
   \resizebox{\linewidth}{!}{
  \begin{tabular}{@{}l|ccc}
    \toprule
    Method  & I-AUROC & AUPRO & P-AUROC\\
    \midrule
    5-shot & 0.834 & 0.936 & 0.984 \\
    10-shot & 0.871 & 0.943 & 0.987 \\
    50-shot & 0.926 & 0.962 & 0.989 \\
    Full dataset  & 0.971 & 0.968 &  0.993  \\
    \bottomrule
  \end{tabular}}
  \vspace{-2mm}
     \caption{Performance of LSFA under few-shot settings.}
  \label{tab:few-shot}
  
\end{minipage}
\end{minipage}
\begin{minipage}{0.34\linewidth}

   \small
  \resizebox{\linewidth}{!}{
\begin{tabular}{c|c|c|c|c} \toprule  
Metric     & S-1    & S-2     & S-3   & All      \\ \midrule
I-AUROC    &95.42   &94.26     &92.14    &84.57    \\ \midrule
AUPRO      &96.01   &95.45     &95.21    &90.15    \\ \bottomrule
\end{tabular}}
\vspace{-2mm}
  \caption{\small Different fine-tuning schemes for RGB+3D on MVTec-3D. 'S-N'/'All' denotes training last N blocks/the whole network.. 
  }

\label{tab:finetuning}

\end{minipage}
 \end{table*}

\begin{table}
\scriptsize
\resizebox{\linewidth}{!}{\begin{tabular}{c|c|c|c|c|c|c} \toprule
             & \multicolumn{3}{c|}{I-AUROC}& \multicolumn{3}{c}{AUPRO}    \\ \toprule  
Method     & 3D    & RGB     & RGB+3D   & 3D    & RGB   & RGB+3D  \\ \midrule
LSFA-LoRA    & 91.06 & 85.43   & 93.91  & 92.17 & 93.97 & 95.16  \\ \midrule
LSFA-AdaLoRA & 91.11 & 85.72   & 93.98  & 92.24 & 94.15 & 95.33 \\ \bottomrule
\end{tabular}}
 \vspace{-2mm}
\caption{\small Training LSFA with LoRA/AdaLoRA on MVTec-3D.}
\label{tab:PEFT}
 \vspace{-1mm}
\end{table}
\noindent\textbf{Investigation on IFC.} We first conduct studies to analyze the influence of the proposed IFC. The method that utilizes pre-trained features without adaptation for PatchCore is used as the baseline for all the evaluations. As shown in Tab.~\ref{tab:ablation_componentall}, the baseline method achieves inferior accuracy for all the metrics with the fixed pre-trained features. By contrast, IFC brings a significant performance boost (about 2.8\%/1.0\%$\uparrow$ for I-AUROC/AUPRO) by explicitly optimizing the feature compactness and keeping consistent with the inference process, which enhances the feature sensitivity to abnormal patterns. Tab.~\ref{tab:ablation_componentIFC} shows a detailed analysis of each loss term within IFC, where both global and local compactness losses contribute to the final performance as well.\\
\textbf{Investigation on CLC.} We then investigate the influence of CLC. As shown in Tab.~\ref{tab:ablation_componentall}, CLC also achieves similar accuracy to IFC by performing multi-granularity cross-modal contrastive representation learning. This mainly accounts for that the proposed CLC can effectively alleviate the impact of inter-modal misalignment from multiple views and meanwhile utilize the informative self-supervised signals for feature extraction. Similarly, Tab.~\ref{tab:ablation_componentCLC} shows the results of each sub-component in CLC, where both global and local cross-modal contrastive losses boost performance over the baseline method. Moreover, further improvement in accuracy can be observed by combining IFC and CLC. Therefore, the above results verified the effectiveness of the proposed IFC, CLC, as well as their own key components.   \\
\textbf{Qualitative results.} We further conduct qualitative experiments to investigate the impact of RGB/3D modality. Fig.~\ref{fig:ablationmemorybank} shows the prediction results of single/combined-modality results. It can be observed that the results of RGB modality are more dispersed and impose large scores in the edge regions as well. By contrast, the distribution of scores for 3D modality is more focused around the defects. Finally, the maps of combining two modalities demonstrate that both two modality helps precise defect localization. \\
\textbf{Parameter sensitivity.} Next, we evaluate the parameter sensitivity of several important hyper-parameters in our method, including the size of the memory bank $n_{I}^{L}$ and the balancing factor $\lambda$. As shown in Fig.~\ref{fig:ablationparameter}~(left), LSFA achieves similar performance across all the sizes, thus not sensitive to $n_{I}^{L}$. To balance the performance and memory cost, we set $n_{I}^{L}=5\times 10^4$. For the influence of $\lambda$, the results in Fig.~\ref{fig:ablationparameter}~(right) demonstrate that LSFA is not sensitive to the value of $\lambda$ as well. Since larger $\lambda$ leads to a slight performance drop, we set $\lambda=0.6$ to get the best results. \\
\textbf{Investigation on adaptor structure.} As shown in Tab.~\ref{tab:structure}, besides the above experiments, we finally investigate the influence of different adaptor structures, including linear projection layer, single vanilla transformer encoder layer, multiple vanilla transformer encoder layers, and 1$\times$1 convolution layer, where the single vanilla transformer encoder layer performs best among these structures.

\subsection{Few-shot Anomaly Detection}
As shown in Tab.~\ref{tab:few-shot}, to evaluate the effectiveness of our method in extreme cases, we conduct experiments on few-shot settings. Specifically, we randomly sample 5/10/50 images from each class as the training set and perform the evaluation on the whole test set. The results show that our method can also achieve superior performance, even compared with some of the methods trained with the whole training set in Tab.~\ref{tab:iaucroc}.  different fine-tuning methods (i.e., LoRA~\cite{lora}) are available as well.

\subsection{Comparison with Fine-tuning Methods}
Here we remove the adaptors $\phi_{I}/\phi_{P}$ and combine LSFA with off-the-shelf fine-tuning methods LoRA~\cite{lora} and AdaLoRA~\cite{adalora} in PEFT. The results are shown in Table.~\ref{tab:PEFT}, which are slightly inferior to results of our LSFA. We remove the adaptors and evaluate the results of training the whole network and training the last few stages of the backbone network in our LSFA respectively. As shown in Table.~\ref{tab:finetuning}, with more modules used for training, a more severe performance drop is observed, especially for training all the blocks. Such phenomenon indicates that training with only part/none of the modules fixed will result in severe catastrophic forgetting and over-fitting to specific data domains, thus failing to distinguish anomalies from normal patterns.

\section{Conclusion}
In this paper, we propose LSFA, a simple yet effective self-supervised multimodal feature adaptation framework for multi-modal anomaly detection. Specifically, LSFA performs feature adaptation in both intra-modal and inter-modal aspects. For the former, a dynamic-updated memory-bank based feature compactness optimization scheme is proposed to enhance the feature sensitivity to unusual patterns. For the latter, a local-to-global consistency alignment strategy is proposed for multi-scale inter-modality information interaction. Extensive experiments show that our method achieves much superior performance than previous methods and meanwhile prominently boosts existing feature embedding based baselines.

{\small
\bibliographystyle{ieee_fullname}
\bibliography{CVPR24}
}

\end{document}